\pdfoutput=1

\documentclass[11pt]{article}

\usepackage[]{EMNLP2023}

\usepackage{times}
\usepackage{latexsym}

\usepackage[T1]{fontenc}

\usepackage[utf8]{inputenc}

\usepackage{microtype}

\usepackage{inconsolata}

\usepackage{svg}
\usepackage{tabularx}
\usepackage{booktabs}
\usepackage{boldline}
\usepackage{multirow}
\usepackage{float}
\usepackage{xcolor}
\usepackage{graphicx}
\usepackage{dialogue}
\usepackage{csquotes}
\usepackage{listings}
\usepackage[para]{threeparttable}
\usepackage{soul} 
\usepackage{enumitem}
\usepackage{mathtools}
\usepackage{adjustbox}
\usepackage{makecell}
\usepackage{amsfonts}
\usepackage{colortbl}
\usepackage{color}
\usepackage[T1]{fontenc}
\usepackage{xspace}
\usepackage{pifont}

\usepackage{CJKutf8} 

\newcommand{\ours}{$\texttt{RER}$\xspace}

\definecolor{grayout}{gray}{0.9}

\title{Rethinking the Role of Proxy Rewards in Language Model Alignment}

\author{
    \textbf{Sungdong Kim}$^{1,2,3}$ \quad \quad \textbf{Minjoon Seo}$^{3}$ \\
    NAVER Cloud$^{1}$ \quad NAVER AI Lab$^{2}$ \quad KAIST AI$^{3}$ \\
    \texttt{sungdong.kim@navercorp.com} \quad \texttt{minjoon@kaist.ac.kr}
}

\begin{document}
\maketitle

\begin{abstract}
Learning from human feedback via proxy reward modeling has been studied to align Large Language Models (LLMs) with human values. However, achieving reliable training through that proxy reward model (RM) is not a trivial problem, and its behavior remained as a black-box. In this paper, we study the role of proxy rewards in the LLM alignment via `reverse reward engineering' by composing interpretable features as a white-box reward function. We aim to replicate the ground truth (gold) reward signal by achieving a \textit{monotonic} relationship between the proxy and gold reward signals after training the model using the proxy reward in reinforcement learning (RL). Our findings indicate that successfully emulating the gold reward requires generating responses that are relevant with enough length to open-ended questions, while also ensuring response consistency in closed-ended questions. Furthermore, resulting models optimizing our devised white-box reward show competitive performances with strong open-source RMs in alignment benchmarks. We highlight its potential usage as a simple but strong reward baseline for the LLM alignment, not requiring explicit pairwise human feedback dataset and RM training~\footnote{Our code is available at \href{https://github.com/naver-ai/rethinking-proxy-reward}{github.com/naver-ai/rethinking-proxy-reward}.}.

\end{abstract}

\section{Introduction}

To align large language models (LLM) with human values like helpfulness, human feedback-based learning has been studied~\cite{ouyang2022training, bai2022training, bai2022constitutional, rafailov2023direct}. Reinforcement Learning from Human Feedback (RLHF) is a dominant approach to exploiting human feedback. Typically, the human feedback is used to train a proxy reward model (RM), and a policy model is optimized over the reward signal from the RM using RL~\cite{schulman2017proximal, ziegler2020finetuning}.

\begin{figure}[t] 
\centering
\includegraphics[width=0.48\textwidth]{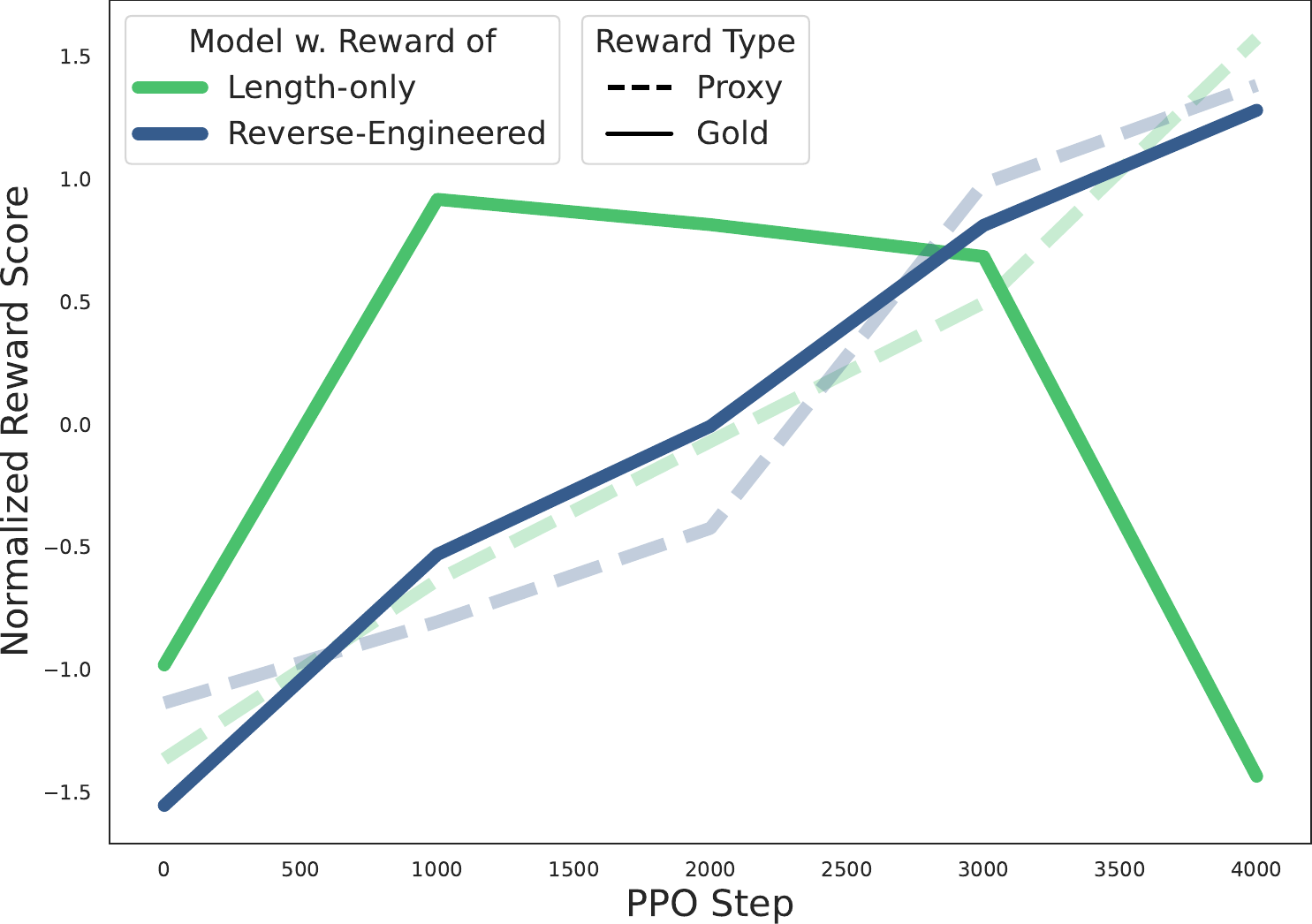}
\caption{A preview of our reverse reward engineering experiment. First, we design white-box reward functions with interpretable features such as the length or relevance of the response. Then, we conduct RL training using each of the designed functions as a proxy reward and deem it a success in reverse engineering if a \textit{monotonic} relationship between the proxy and the ground truth (Gold) reward scores is observed in the multiple evaluations. The reverse-engineered reward (blue) exhibits such a tendency, whereas the length-only reward (green) does not achieve the monotonic relationship, showing reward overoptimization~\cite{gao2023scaling}.}
\label{fig:overview}
\vspace{-2mm}
\end{figure}

However, reliable RL training with the proxy RM is not a trivial problem. \citet{gao2023scaling} study the overoptimization problem of the policy training against the imperfect RM. It is a phenomenon that the proxy rewards from the RM are increased continuously, but the true rewards are saturated or even decreased in fact. Similarly, the policy model often finds undesired shortcuts in the reward signal from the imperfect black-box RM, i.e., ``reward hacking''~\cite{skalse2022defining, pang2022reward}.

A line of research has studied the innate limitations of human feedback. \citet{xu2023critical} reveal prior knowledge affects the preference judgment in that crowd workers often choose the preferred response considering surface-level properties such as conciseness and specificity, while experts focus on more essential properties like factuality and completeness. Similarly, \citet{hosking2023human} further categorize the attributions of the preference judgments and confirm that human judgments could be biased by stylistic features like assertiveness. Moreover, \citet{singhal2023long} analyze the correlations between long response and RLHF, demonstrating a verbosity bias in the human feedback.

In this work, we show merely combining interpretable features as a proxy reward function can substantially maximize the score of state-of-the-art RM~\cite{lambert2024rewardbench} and further enhance human preferences in various alignment benchmarks~\cite{vicuna2023, alpaca_eval, zheng2023judging}. Specifically, we study the role of proxy rewards learned by human feedback in the RLHF by employing ``reverse reward engineering'' with interpretable features, such as length or relevance of the responses, to compose the white-box reward function, as shown in Figure~\ref{fig:overview}. We define the goal of reverse engineering as achieving a \textit{monotonic} relationship between the proxy and ground truth (gold) reward signals after training the model with the proxy reward in an RL manner. This is to verify whether training to maximize the proxy white-box reward during RL training also maximizes the gold reward signal at the test time.

Experimental results indicate that the key to imitating the gold reward signal lies in generating responses that are sufficiently lengthy yet relevant and faithful for open-ended queries while ensuring consistency in responses to closed-ended queries. Contrary to previous study~\cite{singhal2023long}, our results demonstrate that solely optimizing towards lengthy response makes drastic drops in the gold reward, showing severe overoptimization. Also, the reward branching according to query type reliably improves the win rate while not introducing unnecessary verbosity compared to other baseline reward options. Furthermore, we find that the reward differentiation brings advantages in improving preference while minimizing alignment tax~\cite{ouyang2022training}, a phenomenon where the increased preference accompanies the degraded performance on other NLP tasks. Notably, the engineered white-box reward often results in even more improvements than the strong open-source RM like UltraRM-13B~\cite{cui2023ultrafeedback} and also generally works well across LLM backbones, demonstrating its potential usage as a baseline reward.

Our contributions are summarized in three folds:
\begin{itemize}
    \setlength\itemsep{0em}
  \item We investigate the role of proxy rewards learned by human feedback via reverse reward engineering with interpretable features.
  \item Our results suggest that the key to replicating the gold reward involves producing responses that are not only detailed but also relevant for open-ended queries while maintaining consistency in responses to closed-ended queries.
  \item We highlight the potential usage of the reverse-engineered white-box reward function as a simple but strong reward baseline not requiring a pairwise human feedback dataset.
\end{itemize}

\section{Related Work}

\paragraph{Reinforcement Learning from Human Feedback}

Reinforcement Learning from Human Feedback (RLHF) is the most prevalent approach to leverage the human preference~\cite{ziegler2020finetuning, ouyang2022training, bai2022training, liu-etal-2022-aligning, scheurer2023training}. These approaches include the reward modeling stage to develop a proxy RM predicting the human preference over the response, e.g., typically as a scalar format. Then, the trained proxy RM computes reward scores for the sampled responses in the RL stage, i.e., rollout. Recently, replacing human feedback with AI or synthetic feedback has been studied~\cite{bai2022constitutional, kim2023aligning, sun2023salmon, cui2023ultrafeedback}.  There are attempts to leverage multiple rewards beyond the overall reward~\cite{touvron2023llama2, wu2023fine, rame2023rewarded, jang2023personalized}. \citet{wu2023fine} devise fine-grained rewards consisting of relevance, factuality, and information completeness to enhance long-form question-answering.

\paragraph{Pitfalls of Reward Modeling and Human Feedback}

RLHF faces various challenges, especially in reward modeling and human feedback itself~\cite{casper2023open}. \citet{gao2023scaling} present the overoptimization problem by the imperfect proxy reward models. The issue arises when the policy, tailored to enhance the estimated rewards, fails to improve real-world rewards. Many studies try to mitigate the overoptimization problem~\cite{moskovitz2023confronting, coste2023reward}. Similarly, \citet{shen2023trickle} study reward inconsistency and leverage a sentence encoder to increase the reward consistency. The work leverages SimCSE~\cite{gao2021simcse} to augment the preference dataset and normalize reward scores.
The policy models within the RLHF often find and exploit undesired shortcuts or errors in the RMs to maximize the proxy rewards. The phenomenon is known as reward hacking~\cite{amodei2016concrete, pan2022effects, pang2022reward, song2023reward, eisenstein2023helping}. Much literature points out that human preference can be easily biased by superficial features such as complexity and assertiveness~\cite{xu2023critical, hosking2023human}. \citet{singhal2023long} study a high correlation between response length and RLHF training. Motivated by these studies, we integrate a set of interpretable features, such as length incentive, repetition penalty, query relevance, and distinction by query type, to understand the behaviors of the learned proxy reward models. 

\section{Experimental Setup}

\begin{figure}[t] 
\centering
\includegraphics[width=0.48\textwidth]{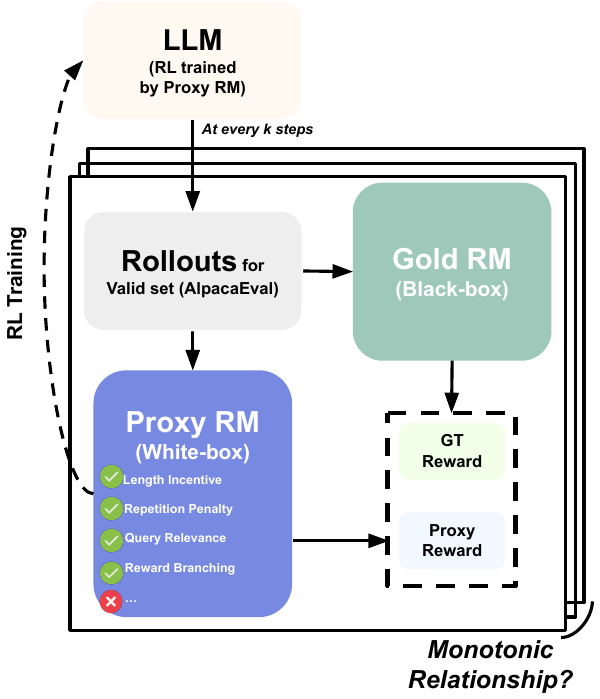}
\caption{An overview of reverse \textit{reward} engineering study. It aims to imitate the ground-truth reward signal by Gold RM with white-box reward features such as length, repetition, and relevance. Specifically, we try to observe the monotonic relationship between the proxy and gold reward signals across the multiple evaluations during RL training. We could comprehend the roles of Gold RM via the interpretable features from the study.}
\label{fig:reverse-overview}
\vspace{-2mm}
\end{figure}

\subsection{Reverse Reward Engineering}

Inspired by previous studies on superficial biases in human feedback~\cite{singhal2023long, xu2023critical, hosking2023human}, we aim to explore whether there would be a reward function with interpretable features that can maximize the ``ground truth'' reward signal in reverse, as shown in Figure~\ref{fig:reverse-overview}. Started by the length bias~\cite{singhal2023long, chen2024odin}, we progressively design the additional features to achieve the goal. More formally, we first perform RL training, i.e., Proximal Policy Optimization (PPO)~\cite{schulman2017proximal}, with the designed function as a proxy reward and then consider it a success in reverse engineering if a \textit{monotonic} relationship between the proxy reward and the ground truth reward is observed in the multiple evaluations over validation set. In other words, if overoptimization~\cite{gao2023scaling} for the proxy reward is detected, it will be considered a failure in reverse engineering.

\subsection{Features for White-box Reward}
\label{subsec:reverse-reward-hacking}

In this subsection, we detail the reward features employed in our experiments on reverse engineering. For a set of $(x, y)$ pairs, we denote $x$ as a query, $y$ as a reference response, and $\hat{y}$ as a rollout response of Proximal Policy Optimization (PPO)~\cite{schulman2017proximal, ziegler2020finetuning}.

\paragraph{Length Incentive (\texttt{LI}).}

First, we introduce Length Incentive (\texttt{LI}) to exploit the verbosity bias in human preference~\cite{singhal2023long, wang2024far, chen2024odin}. It scores the reward by calculating $\texttt{LI}(\hat{y})\ =\ \frac{\text{\# words of}\ \hat{y}}{100}$. Intuitively, this reward function promotes the longer generation regardless of the given query $x$. It is similar to LPPO suggested by ~\cite{singhal2023long}, but we do not normalize the reward with the target length.

\paragraph{Repetition Penalty (\texttt{RP}).}

In a preliminary study, solely using \texttt{LI} as a reward function produces undesired repetitions in the generations~\cite{welleck2019neural}.
We employ the Repetition penalty (\texttt{RP}) to mitigate unnecessary repetitions in $\hat{y}$. It checks the unique trigram ratio of $\hat{y}$, i.e., $\texttt{RP}(\hat{y}) = \frac{\text{\# unique trigram of}\ \hat{y}}{\text{\# trigrams of}\ \hat{y}}$. Integrating it with the Length Incentive function efficiently prevents redundancy and encourages proper length of outputs, i.e., $\texttt{LI}(\hat{y}) \cdot \texttt{RP}(\hat{y})$.

\paragraph{Query Relevance (\texttt{QR}).}

We find that combining \texttt{LI} and \texttt{RP} is somewhat effective, but the functions often promote the generation of irrelevant content since both functions are input-agnostic.
Hence, we involve the Query Relevance (\texttt{QR}) as one of the reward features. It checks whether the generated response $\hat{y}$ contains off-topic contents for a provided query $x$. We compute the relevance score between the query and sampled response: $\texttt{QR}(x, \hat{y}) = M(x) \cdot M(\hat{y}) \in \mathbb{R}^1$, where the $M$ is pre-trained bi-encoder retriever such as Contriever~\cite{izacard2021unsupervised}.

\paragraph{Reward Branching and Reference Answer Relevance (\texttt{AR}).}

The intuition behind the mixture of reward functions $\texttt{LI}(\hat{y}) \cdot \texttt{RP}(\hat{y}) \cdot \texttt{QR}(x, \hat{y})$ is aiming for high relevance between query and generated response while promoting long response length and less repetitions.
However, the reward functions, especially \texttt{LI}, could be problematic when the query requires constrained or factual responses. In these cases, consistency of response might be much more important than the long and diverse responses, as discussed in \citet{touvron2023llama2, song2023reward}.

We apply the different relevance rewards according to query type to handle such cases. Specifically, we define two types of queries, \texttt{Open-ended (OE)} and \texttt{Closed-ended (CE)}, according to whether they require creative and open-ended or consistent and constrained responses, similar to \citet{song2023reward}. Then, we include reference answer relevance $\texttt{AR}(y, \hat{y}) = M(y) \cdot M(\hat{y})$, where the $y$ is the ground-truth reference response of given query. Finally, we denote the final reward design, Reverse Engineered Reward (\ours):

\begin{equation*}
    \texttt{RER} = 
    \begin{cases}
        \texttt{LI}(\hat{y}) \cdot \texttt{RP}(\hat{y}) \cdot \texttt{QR}(x, \hat{y})         & \small{\text{if } T(x) = \texttt{OE}} \\
        \texttt{RP}(\hat{y}) \cdot \mathcal{F}(\texttt{AR}(y, \hat{y}))    & \small{\text{else }} \\
        
    \end{cases}
\end{equation*}

, where the $T(x) \in \{\texttt{OE}, \texttt{CE}\}$ indicates the pre-identified query types and the $\mathcal{F}$ indicates a linear interpolation function mapping the range of \texttt{AR} to the range of $\texttt{LI}(\hat{y})$ for stable RL training~\footnote{We empirically find that the multiplication of each reward function works better than the addition of them. Please see Appendix~\ref{appendix:combine-reward} for more details.}.

\paragraph{Query-Type Classification.}

Deciding whether the given query requires open-ended generation may not be obvious because of its subjectivity in interpreting the range of `open-ended'. Nevertheless, we define the meaning of open-ended so that the responses to the query can be relatively anything and creative. For example, ``How can I make a good first impression?'' could be an open-ended query. On the other side, the closed-ended query indicates the corresponding responses should be consistent and constrained as a closed form. ``How many movies are in the Fast and Furious franchise?'' would be one of the closed-ended queries.
Considering resource constraints, we automatically annotate the query type $T(x)$ (\texttt{OE} or \texttt{CE}) used in \ours by prompting it to GPT-4~\cite{openai2023gpt4}~\footnote{We find that about 83\% of decisions by GPT-4 are agreed upon when the authors manually validate 100 samples.}. Please see Appendix~\ref{appendix:query-type-cls} for more details of the query-type classification. 

\paragraph{PPO training.}

We use the resulting white-box reward function, \ours, to further optimize an SFT model with reinforcement learning. Consequently, the following PPO objective induces the model to maximize \ours,

\begin{equation*}
     \max_{\pi_{\phi}}\mathbb{E}_{\substack{(x,y) \sim D,\\ \hat{y}\sim\pi_{\phi}(\cdot|x)}}[\texttt{RER}(x, \hat{y}, y) - \beta \log \left(\dfrac{\pi_{\phi}(\hat{y}|x)}{\pi_{\rho}(\hat{y}|x)}\right)],
\end{equation*}

while online policy $\pi_{\phi}$ and the fixed reference (SFT) policy $\pi_{\rho}$ do not diverge too much by regularizing it with KL constraints~\cite{schulman2017proximal, ziegler2020finetuning}, where the $D$ is train dataset and the $\beta$ is the coefficient to control the KL penalty.

\subsection{Evaluation setup}

\paragraph{Implementation Details.}

We conduct experiments based on two alignment datasets, Anthropic-HH~\cite{bai2022training} and AlpacaFarm~\cite{dubois2023alpacafarm}. We perform PPO~\cite{ziegler2020finetuning} training with rewards from \ours or other reward models at most 5k steps. As in Section~\ref{subsec:reverse-reward-hacking}, we employ Contriever~\cite{izacard2021unsupervised} for scoring relevance scores~\footnote{\href{https://huggingface.co/facebook/contriever}{huggingface.co/facebook/contriever}}. We train SFT model based on LLaMA-2-7B~\cite{touvron2023llama2} with the 161k chosen responses in Anthropic-HH~\footnote{\href{https://huggingface.co/datasets/Anthropic/hh-rlhf}{huggingface.co/datasets/Anthropic/hh-rlhf}}. For the PPO training on Anthropic-HH, we select about 23k (train and dev) first-turn queries from `Helpful-rejection' and `Helpful-online' sub-splits for efficient implementation.  Also, we employ the pre-trained SFT model for the AlpacaFarm. We use 20k of `unlabeled' split for the PPO training on AlpacaFarm~\footnote{\href{https://huggingface.co/datasets/tatsu-lab/alpaca\_farm/viewer/alpaca\_instructions}{huggingface.co/datasets/tatsu-lab/alpaca\_farm}}. More details are in Appendix~\ref{appendix:implementation}.

\paragraph{Gold RM.}

As discussed in \citet{gao2023scaling}, it is a non-trivial problem to obtain human labels for the ground-truth reward signal. Hence, following the \citet{gao2023scaling}, we also assume the gold-standard reward model (Gold RM) to evaluate the reward functions. Specifically, we employ one of the strong open-source RMs, StarlingRM-34B~\footnote{\href{https://huggingface.co/Nexusflow/Starling-RM-34B}{huggingface.co/Nexusflow/Starling-RM-34B}}, as the Gold RM. It is trained on the Yi-34B-Chat~\cite{young2024yi} with Nectar, which is an AI feedback dataset~\cite{starling2023}. This model demonstrates the best performance on various preference evaluation tasks within the recently proposed RewardBench~\cite{lambert2024rewardbench}. We check the reward signal relationship between proxy and gold rewards based on 805 prompts in AlpacaEval~\cite{alpaca_eval}.

\begin{figure*}[t] 
\centering
\includegraphics[width=\textwidth]{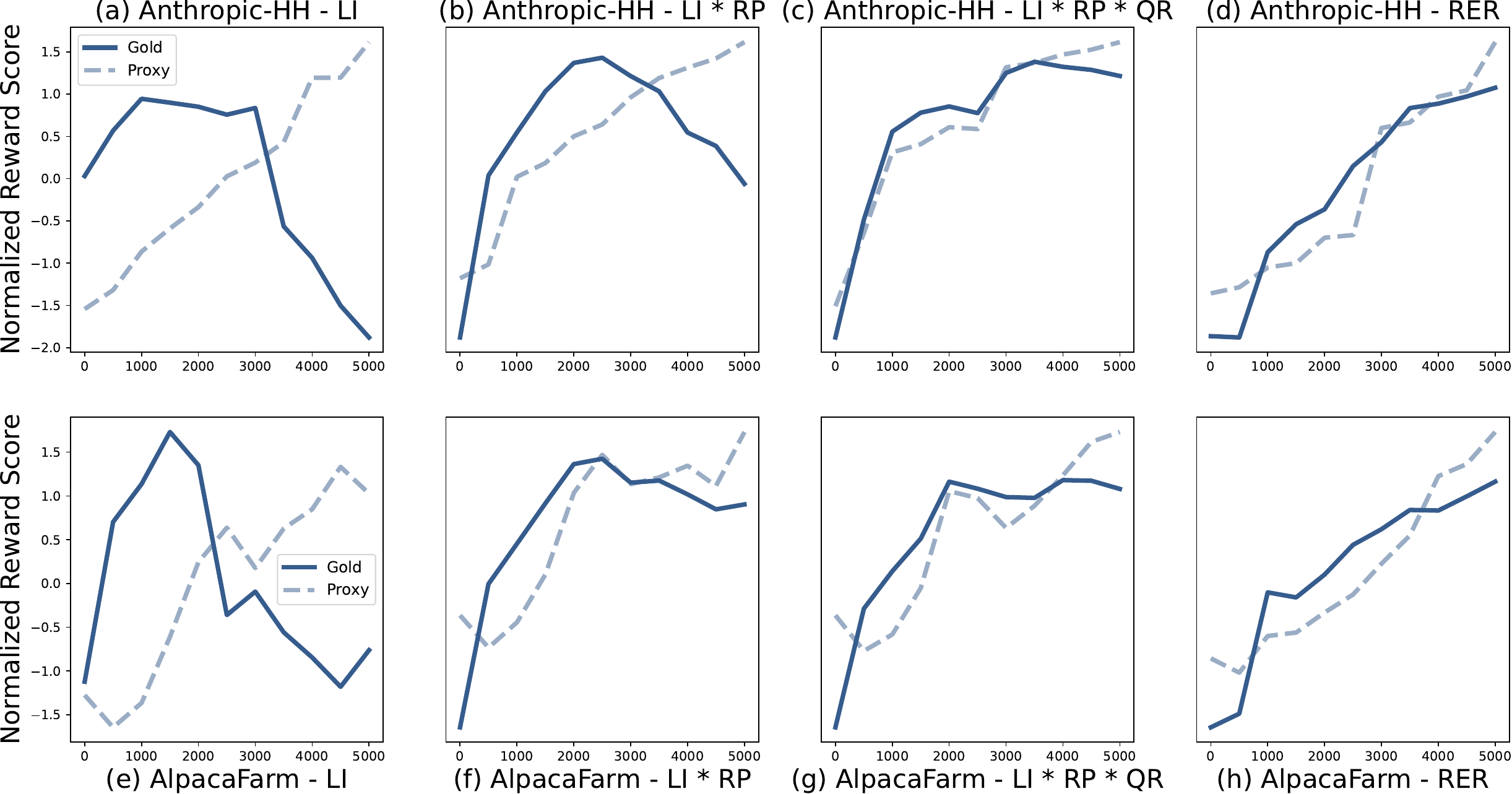}
\caption{Results of reverse reward engineering. We visualize normalized proxy and gold reward scores for every 500 PPO steps against each reward design option. The results on the upper side are from Anthropic-HH~\cite{bai2022training}, and the results on the lower side are from AlpacaFarm~\cite{dubois2023alpacafarm}, respectively. Instances of AlpacaEval~\cite{alpaca_eval} are used to compute the reward scores. We expect a monotonical relationship between the proxy and gold reward scores to achieve success in reverse engineering. We find that considering the relevance and adopting different rewards according to query type, i.e., \ours, contribute to increasing the gold reward reliably.}
\label{fig:hhrlhf-normalized-reward}
\end{figure*}

\paragraph{Other Evaluations.}

We further evaluate the resulting models on \textbf{Vicuna-Bench}~\cite{vicuna2023}, \textbf{AlpacaEval}~\cite{alpaca_eval}, and \textbf{MT-Bench}~\cite{zheng2023judging}. The three benchmarks mainly evaluate the preference of models' responses leveraging a superior proprietary LLM, GPT-4~\cite{openai2023gpt4}. In addition, we include Super-NaturalInstructions (\textbf{SuperNI})~\cite{wang2022super} to investigate the alignment tax problem of the resulting models. More evaluation details are in Appendix~\ref{appendix:evaluation}.

\paragraph{Relevant Sentence Ratio.}

To assess the relevance of the model's responses for the provided query, we measure the Relevant Sentence Ratio (\% Rel. Sent). Specifically, we request GPT-4~\cite{openai2023gpt4} to refer to each sentence first and then judge its query relevance as either `Relevant' or `Irrelevant'. That is, it indicates the ratio of on-topic sentences in a generated response. The regarding prompt for this evaluation is found in Appendix~\ref{appendix:rel-eval}.

\paragraph{Baseline RMs.}

In addition to the Gold RM, i.e., \textbf{StarlingRM-34B}, used for our analysis, we include three open-source RMs as baselines.
\textbf{OpenAssistant (OASST)} is trained with four public feedback datasets, WebGPT~\cite{nakano2021webgpt}, OpenAI Summary~\cite{stiennon2020learning}, Anthropic-HH~\cite{bai2022training}, and SyntheticGPT~\footnote{\href{https://huggingface.co/datasets/Dahoas/synthetic-instruct-gptj-pairwise}{Dahoas/synthetic-instruct-gptj-pairwise}}, based on DeBERTa-V3-Large~\cite{he2021debertav3}. \textbf{SteamSHP-XL} is another open-source RM trained with SHP~\cite{pmlr-v162-ethayarajh22a} and Helpfulness subset of the Anthropic-HH. Unlike other feedback datasets, the SHP consists of human-generated responses from an online community, Reddit. It is based on FLAN-T5-XL~\cite{chung2022scaling}. Also, we include \textbf{UltraRM-13B}, one of the top open-source RMs trained on UltraFeedback~\cite{cui2023ultrafeedback}. The UltraFeedback is a 64k AI feedback dataset constructed by GPT-4~\cite{openai2023gpt4}. Also, it is further fine-tuned with the public feedback datasets, Anthropic-HH, SHP, and OpenAI Summary. We perform PPO training against the reward score by each RM or our reward functions.

\section{Experiments}

\subsection{Reverse Reward Engineering}
\label{subsec:reverse-reward-hacking-exp}

We compare four combinations of the features as a proxy reward function, (1) \texttt{LI}, (2) $\texttt{LI} \cdot \texttt{RP}$, (3) $\texttt{LI} \cdot \texttt{RP} \cdot \texttt{QR}$, and (4) \ours. To capture the monotonic relationship between proxy and both reward signals, we compute reward scores of both proxy and gold for AlpacaEval~\cite{alpaca_eval} set at every 500 PPO steps against the proxy reward.
Figure~\ref{fig:hhrlhf-normalized-reward} and Table~\ref{table:correlation} show the results for the reverse reward engineering based on the Anthropic-HH~\cite{bai2022training} and AlpacaFarm~\cite{dubois2023alpacafarm} datasets. We perform standard normalization over the reward scores for the plotting.

\textbf{We find that the solely promoting lengthy response, i.e., \texttt{LI} as a proxy reward, fails to monotonically increase the gold reward signal,} contrary to recent findings in \citet{singhal2023long}. More specifically, it improves the gold reward for the initial 1-1.5k steps but subsequently leads to a consistent decrease in the gold reward for the remaining steps even though the proxy rewards are consistently improved, as shown in Figure~\ref{fig:hhrlhf-normalized-reward} (a) and (e). Recalling that the PPO steps were set to at most 400 in the \citet{singhal2023long}, we presume that the \texttt{LI} is effective only for a few initial steps. Penalizing unnecessary repetitions along with lengthy responses shows a better tendency, yet it eventually results in a decrease in the gold reward (Figure~\ref{fig:hhrlhf-normalized-reward} (b) and (f)). However, the magnitude of this decrease is considerably less than that experienced with \texttt{LI} alone. The \texttt{LI} is the only reward that exhibits negative correlations with the gold reward signal in Table~\ref{table:correlation}.

\textbf{Considering relevance along with the features reliably increases the gold reward, indicating the success of reverse engineering.} We find the \texttt{QR} contribute to avoiding the drastic drop in gold reward score and even achieves the success of the reverse engineering, improving the SpearmanR up to 0.86, as shown in Figure~\ref{fig:hhrlhf-normalized-reward} (c), (g) and Table~\ref{table:correlation}. On the other hand, \ours monotonically increases the gold reward signal for both datasets, as illustrated in Figure~\ref{fig:hhrlhf-normalized-reward} (d) and (h). Also, in Table~\ref{table:correlation}, \ours achieves the highest Spearman correlations, indicating a strong monotonic relationship with the gold reward signal on both datasets. Although \ours shows a relatively slower start compared to other reward options, it exhibits the most consistent upward trend. However, we find the relevance alone, e.g., \texttt{QR}, can not improve the gold reward while showing a similar tendency with the \texttt{LI}. It implies that aiming for sufficiently lengthy and faithful responses while avoiding off-topic or unnecessary repetitions is key to maximizing the gold reward signal. These results have a connection with the previous findings on superficial biases like assertiveness in human preference~\cite{xu2023critical, hosking2023human}.

\begin{figure}[t] 
\centering
\includegraphics[width=\columnwidth]{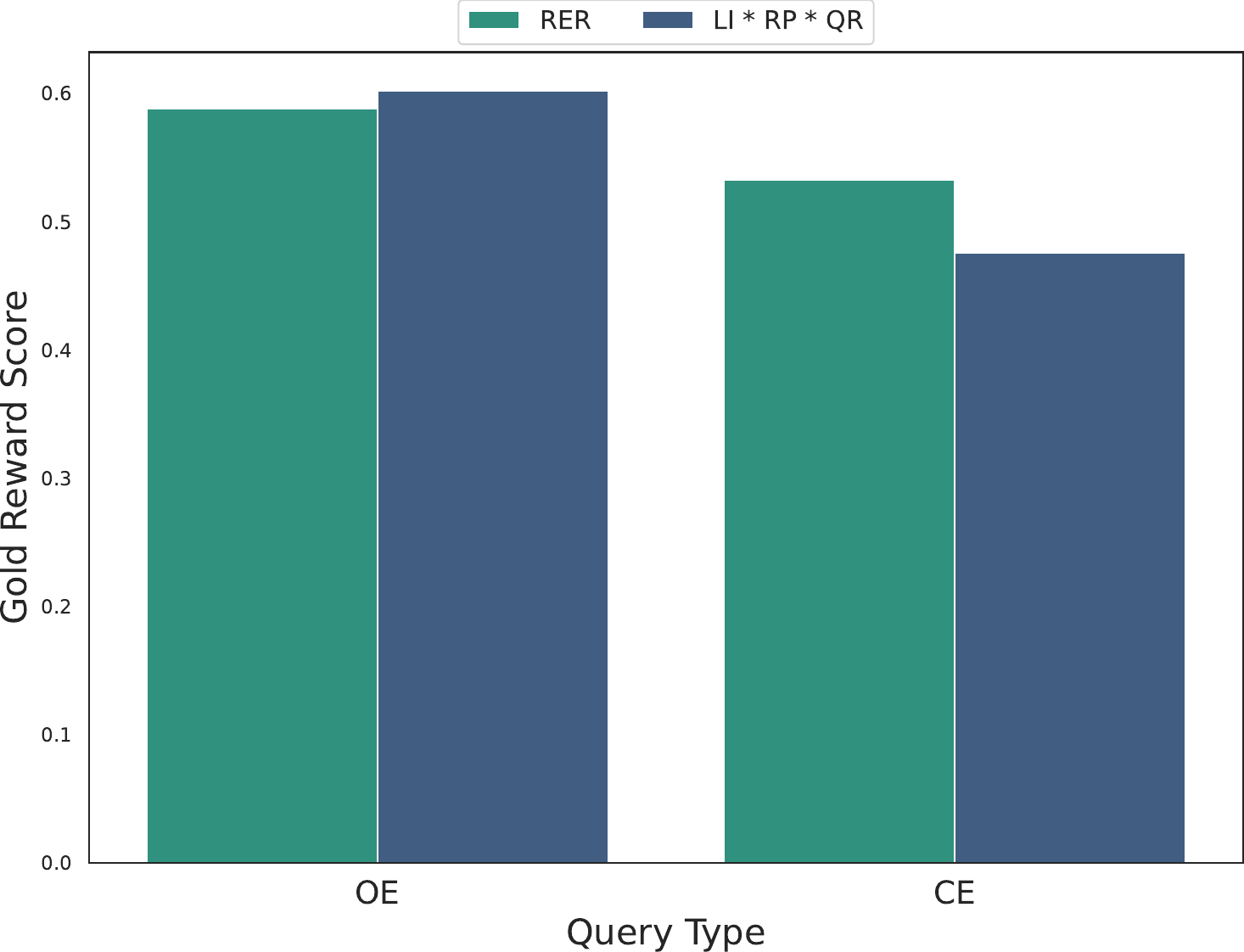}
\caption{Gold reward scores according to whether query type requires open-ended (\texttt{OE}) or closed-ended (\texttt{CE}) responses. We compare two proxy reward options, $\texttt{LI} \cdot \texttt{RP} \cdot \texttt{QR}$ and \ours, based on models trained with Anthropic-HH~\cite{bai2022training}. We find they show meaningful differences in \texttt{CE} type, demonstrating the importance of the reward branching.}
\label{fig:gold-reward-score-by-qtype}
\end{figure}

\textbf{Reward branching according to whether the query requires open-ended responses makes a meaningful difference, especially for \texttt{CE} type queries.} We compare the final policies optimized and selected by proxy reward functions. Specifically, we measure reward scores using the Gold RM based on the AlpacaEval set, as reported in Figure~\ref{fig:gold-reward-score-by-qtype}. We find the reward scores are decreased if the model exploits  \texttt{LI} regardless of the query type, i.e., no reward branching. We will show this gap makes further differences in the alignment tax problem~\cite{askell2021general, ouyang2022training}, which is a phenomenon showing performance drops in other NLP tasks after alignment, in Section~\ref{subsec:exp-alignment-tax}.

\begin{figure}[t] 
\centering
\includegraphics[width=\columnwidth]{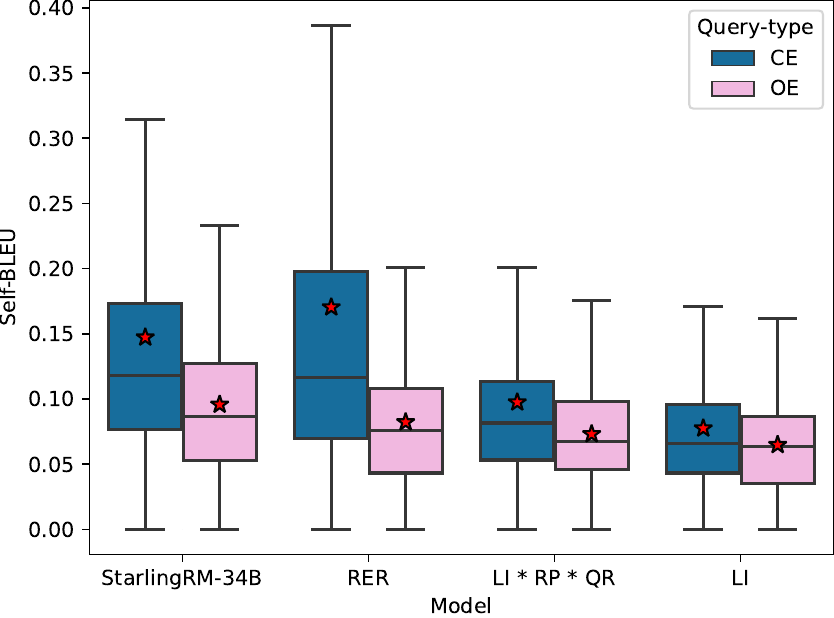}
\caption{Self-BLEU of the PPO models on AlpacaEval according to query types. Please note that the higher Self-BLEU indicates lower diversity for the responses, i.e., consistent responses, and vice versa. As described in \citet{touvron2023llama2}, we intend the high Self-BLEU for the \texttt{CE} type and low Self-BLEU for the \texttt{OE} type.}
\label{fig:self-bleu}
\end{figure}

Additionally, we compare the models' diversity and consistency according to the query type in Figure~\ref{fig:self-bleu}. We report the Self-BLEU~\cite{zhu2018texygen} by sampling 10 responses for each input in the AlpacaEval~\footnote{We perform nucleus sampling~\cite{holtzman2019curious}.}. A high Self-BLEU score indicates a consistent response, whereas a low Self-BLEU score denotes a diverse response~\cite{touvron2023llama2}. As we intended, \ours improves the consistency in the \texttt{CE} type queries and promotes more diverse responses in the \texttt{OE} type, showing a similar tendency to that of Gold RM. However, other models without reward branching produce excessively diverse responses regardless of the query type.

\begin{table}[t!]
    \centering
    \small
    \adjustbox{max width=\columnwidth}{%
    \begin{tabular}{lcccc}
        \toprule
        \textbf{Reward} & \multicolumn{2}{c}{\textbf{Anthropic-HH}} & \multicolumn{2}{c}{\textbf{AlpacaFarm}} \\
        & \small{PearsonR} & \small{SpearmanR} & \small{PearsonR} & \small{SpearmanR} \\
        \midrule
        \textit{w.} \texttt{LI} & -0.77 & -0.66 & -0.52 & -0.58  \\
        \textit{w.} $\texttt{LI} \cdot \texttt{RP}$ & 0.52 & 0.05 & 0.70 & 0.63 \\
        \textit{w.} $\texttt{LI} \cdot \texttt{RP} \cdot \texttt{QR}$ & \textbf{0.96} & 0.86 & 0.78 & 0.86 \\
        \textit{w.} \ours & 0.92 & \textbf{0.99} & \textbf{0.86} & \textbf{0.97} \\
        \bottomrule
    \end{tabular}}
    \caption{We report Pearson and Spearman correlation (R) between proxy and gold reward scores across multiple evaluations. For both datasets, \texttt{RER} shows a SpearmanR close to 1. This indicates a \textit{monotonic} relationship between the two reward signals, signifying successful reverse reward engineering.}
    \label{table:correlation}

\end{table}

\subsection{Comparison with other open-source RMs}
\label{subsec:exp-hhrlhf}

We compare our reward designs with open-source RMs in improving human preference, e.g., producing more helpful responses, based on the Anthropic-HH dataset~\cite{bai2022training}.

\begin{table}[t]
    \centering
    \small
    \adjustbox{max width=\columnwidth}{%
    \begin{tabular}{lccc}
        \toprule
        \makecell{\textbf{Model}\\ \textbf{ \small{(Anthropic-HH)}}}  & \makecell{\textbf{Vicuna} \\\textbf{Bench}} & \makecell{\textbf{Alpaca} \\\textbf{Eval}} & \makecell{\textbf{MT} \\ \textbf{Bench}} \\
        \midrule
        SFT & 50.0 & 15.9 & 3.9 \\
        \midrule
        \textit{+ PPO} \\
        \textit{w.} SteamSHP-XL & 59.4 & 19.6 & 3.8 \\
        \textit{w.} OASST & 61.3 & 19.1 & 3.9 \\
        \textit{w.} UltraRM-13B & 73.8 & \underline{28.0} & 4.2 \\
        \textit{w.} StarlingRM-34B & \textbf{79.4} & \textbf{29.1} & \underline{4.3} \\
        \midrule
        \textit{w.} \texttt{LI} & 67.5 & 16.9 & 3.4  \\
        \textit{w.} $\texttt{LI} \cdot \texttt{RP}$ & 71.9 & 20.3 & 3.9 \\
        \textit{w.} $\texttt{LI} \cdot \texttt{RP} \cdot \texttt{QR}$ & 71.9 & 23.2 & 4.0 \\
        \textit{w.} \ours & \underline{76.9} & 23.4 & \textbf{4.4} \\
        \bottomrule
    \end{tabular}}
    \caption{Comparison of the designed rewards with open-source RMs trained on human or AI feedback, based on Anthropic-HH~\cite{bai2022training}. The PPO model optimizing \ours shows competitive performances with models trained with open-source RMs.}
    \label{table:hhrlhf-exp}
\end{table}

\textbf{Optimizing only towards lengthy responses, i.e., \texttt{LI}, still improves the alignment scores compared to the SFT baseline, even if it shows drastic drops in Gold RM score as in Section~\ref{subsec:reverse-reward-hacking-exp}.} In Table~\ref{table:hhrlhf-exp}, the PPO$_\texttt{LI}$ model even performs better SteamSHP-XL and OASST based PPO models in Vicuna-Bench~\cite{vicuna2023}. This result reminds us of the existence of verbosity bias in LLM-as-a-judge evaluation~\cite{zheng2023judging}. On the other hand, the contributions of \texttt{RP}, \texttt{QR}, and reward branching are shown as a similar pattern with the results of Section~\ref{subsec:reverse-reward-hacking-exp}. Notably, the PPO model trained with \ours achieves even better performances than the model trained with UltraRM-13B in Vicuna-Bench and MT-Bench.

\begin{table}[t]
    \centering
    \adjustbox{max width=\columnwidth}{%
    \begin{tabular}{lc|ccc}
        \toprule
        \makecell{\textbf{Model}\\ \textbf{ \small{(Anthropic-HH)}}} & \makecell{\textbf{\% Win} \\\textbf{Rate}} & \makecell{\textbf{Avg} \\ \textbf{\# Tokens}} & \makecell{\textbf{4-gram} \\\textbf{Rep.}} & \makecell{\textbf{\% Rel.} \\ \textbf{Sent}} \\
        \midrule
        SFT  & 15.9 & 126 & \textbf{0.05}  & \textbf{85.9} \\
        \midrule
        \textit{+ PPO} \\
        \textit{w.} SteamSHP-XL & 19.6 & 154 & 0.10 & 80.2 \\
        \textit{w.} OAAST & 19.1 & 186 & 0.10 & 85.0 \\
        \textit{w.} UltraRM-13B & \underline{28.0} & 213 & 0.11 & 83.1 \\
        \textit{w.} StarlingRM-34B & \textbf{29.1} & 220 & \underline{0.08} & \underline{85.3} \\
        \midrule
        \textit{w.} \texttt{LI} & 16.9 & 351 & 0.32 & 76.5 \\
        \textit{w.} $\texttt{LI} \cdot \texttt{RP}$ & 20.3 & 285 & 0.17 & 81.6 \\
        \textit{w.} $\texttt{LI} \cdot \texttt{RP} \cdot \texttt{QR}$ & 23.2 & 305 & 0.13 & 84.8 \\
        \textit{w.} \ours & 23.4 & 243 & 0.09 & 85.1 \\
        \bottomrule
    \end{tabular}}
    \caption{We analyze the responses from AlpacaEval~\cite{alpaca_eval}. We report the relevant sentence ratio (\% Rel. Sent), leveraging GPT-4~\cite{openai2023gpt4}. We also reference the number of average tokens (\# Avg Tokens) and 4-gram repetitions (4-gram Rep.). \ours archives the win rate while not increasing unnecessary verbosity.}
    \label{table:hhrlhf-anal}
    \vspace{-1mm}
\end{table}

\textbf{PPO model with \ours achieves competitive performance while showing a response distribution similar to that of the model with the Gold RM in terms of length, repetitions, and relevance.} In Table~\ref{table:hhrlhf-anal}, we report the average number of tokens (Avg \#Tokens), 4-gram Repetitions (Rep.), and Relevant Sentence Ratio (\% Rel. Sent) of models' responses based on the AlpacaEval set. We can see the obvious differences in response distribution among \ours and other designed reward functions. In particular, it exhibits the importance of reward branching in that it achieves the win rate without unnecessary verbosity, i.e., much less response length and repetition while containing more on-topic content (the higher \% Rel. Sent). Qualitative examples among the resulting models are in Appendix~\ref{appendix:qualitative-relevancy-analysis}.

\begin{table}[t]
    \centering
    \adjustbox{max width=\columnwidth}{%
    \begin{tabular}{lcc}
        \toprule
        \makecell{\textbf{Model}\\ \textbf{ \small{(AlpacaFarm)}}} & \makecell{\textbf{AlpacaEval} \\ \small{\% Win / \% Rel}} & \makecell{\textbf{SuperNI} \\ \small{ROUGE-L}} \\
        \midrule
        SFT & 23.2 / \textbf{88.2} & \textbf{34.9}  \\
        \midrule
        \textit{+ PPO} \\
        \textit{w.} UltraRM-13B & 34.1 / 85.1 & 28.8 \\
        \textit{w.} StarlingRM-34B & \textbf{44.0} / \underline{87.3} & 17.3 \\
        \midrule
        \textit{w.} \texttt{LI} & 24.1 / 64.5 & 7.4 \\
        \textit{w.} $\texttt{LI} \cdot \texttt{RP}$ & \underline{41.6} / 81.3 & 13.1 \\
        \textit{w.} $\texttt{LI} \cdot \texttt{RP} \cdot \texttt{QR}$ & 41.5 / 84.8 & 22.6  \\
        \textit{w.} \ours & 37.2 / 86.4 & \textbf{34.9} \\
        \midrule
        \quad \textit{w.} \textit{random} $T(x)$ & 34.8 / 82.6 & 28.4 \\
        \bottomrule
    \end{tabular}}
    \caption{Alignment tax measurement of models from AlpacaFarm~\cite{dubois2023alpacafarm} We include SuperNI~\cite{wang2022super} as a test set to measure how the PPO models retain the instruction-following ability, i.e., zero-shot NLP tasks. The \textit{random} $T(x)$ indicates the query type $T(x)$ is obtained by random.}
    \label{table:alpaca-exp}
\end{table}

\subsection{Investigating Alignment Tax}
\label{subsec:exp-alignment-tax}

We further analyze `\textit{alignment tax}' from the models trained with our devised rewards. The alignment tax is a phenomenon in which the improved preference accompanies degraded performances on other NLP tasks after alignment procedure~\cite{askell2021general, bai2022training, ouyang2022training}. We conduct these experiments based on another alignment dataset, AlpacaFarm~\cite{dubois2023alpacafarm}. Specifically, we evaluate models on SuperNI~\cite{wang2022super}, the collection of zero-shot NLP tasks, to measure alignment tax.

\paragraph{Reward branching and relevance contribute to minimizing alignment tax.}

As shown in Table~\ref{table:alpaca-exp}, \ours reliably improves the preference (\% Win) on the AlpacaEval while keeping high relevance (\% Rel), zero-shot instruction-following ability (SuperNI). If we use random query type, \textit{random} $T(x)$, for the \ours instead of the predicted one by GPT-4, its win rate and relevant sentence ratio are degraded on the AlpacaEval. Also, its instruction-following ability is sacrificed, indicating that reward differentiation based on query types is effective in mitigating alignment tax.

\textbf{There is an observable pattern that clearly exacerbates the alignment tax when excluding each component from \ours.} If we do not use reference answer relevance according to query type, i.e., $\texttt{LI} \cdot \texttt{RP} \cdot \texttt{QR}$, it shows even higher preference scores (\% Win) with decreased relevance (\% Rel). However, the achievement accompanies significant performance drops on the SuperNI by about 12.3\% points. If the \texttt{QR} is discarded ($\texttt{LI} \cdot \texttt{RP}$), the performance of the relevant sentence ratio and ROUGE-L score of SuperNI drop more drastically. Finally, using only the response length as a reward (\texttt{LI}) improves the preference slightly compared to SFT, as reported in \citet{singhal2023long}, but it significantly worsens the \% Rel and SuperNI score in return. On the other hand, while StarlingRM-34B significantly improves the win rate, it notably degrades performance in the SuperNI compared to the SFT baseline, exhibiting a severe alignment tax.

\begin{figure}[t] 
\centering
\includegraphics[width=\columnwidth]{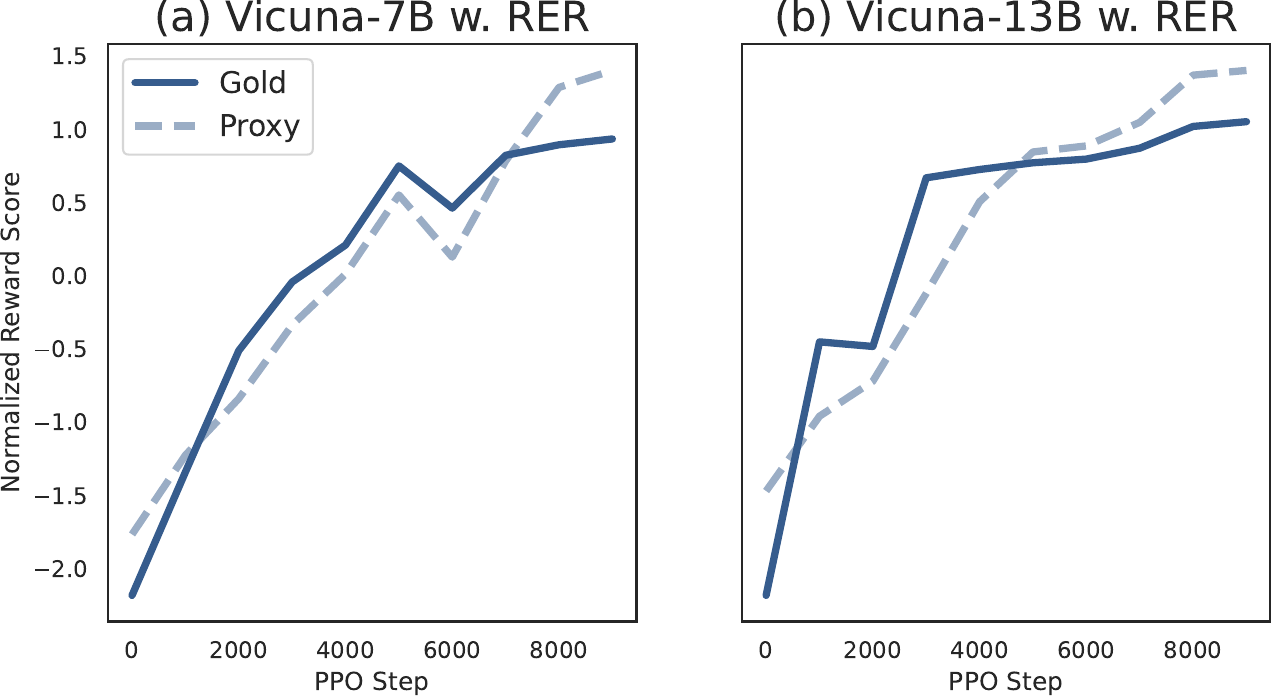}
\caption{Normalized proxy and gold reward scores for every 1k PPO steps against \ours. We conduct the experiments based on two strong SFT models, Vicuna-\{7, 13\}B-v1.5~\cite{vicuna2023, zheng2023judging}, to observe scaling patterns. Both backbones optimizing \ours generally show the monotonic relationship between the \ours and Gold RM scores.}
\label{fig:vicuna-reward}
\vspace{-2mm}
\end{figure}

\subsection{Generalizability of \ours}
\label{subsec:exp-generalizability}

We investigate how \ours is generalizable regarding (1) the number of PPO steps and (2) backbone model size. Specifically, we conduct the experiments based on strong SFT models, Vicuna-\{7, 13\}B-v1.5~\cite{vicuna2023, zheng2023judging}. We sample datasets from ShareGPT~\cite{vicuna2023} for our PPO training. More training details are in Appendix~\ref{appendix:implementation}.

\begin{figure}[t] 
\centering
\includegraphics[width=\columnwidth]{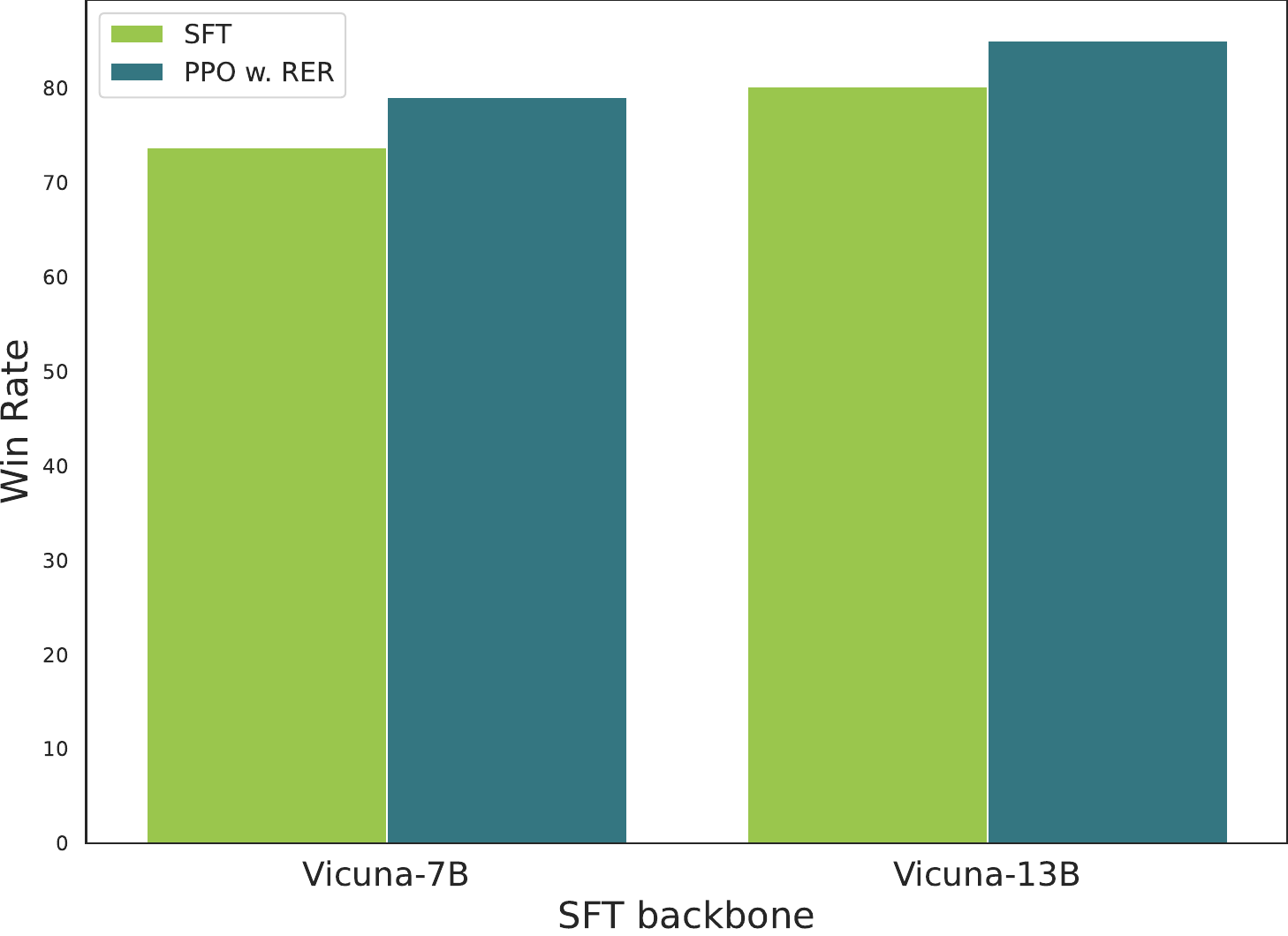}
\caption{Win rate of models trained with \ours compared to their baseline SFT backbones in AlpacaEval~\cite{alpaca_eval}. They show consistent performance gains.}
\label{fig:sft-backbone}
\end{figure}

\textbf{We find the monotonic relationship is consistently observed for both backbone sizes even if we extend the number of PPO steps from 5k to 10k, as shown in Figure~\ref{fig:vicuna-reward}}. However, the upward tendency is saturated after the 5k step and often exhibits fluctuation in the reward scores. It appears that such saturation occurs earlier (2k) in the 13B model. Additionally, we observe that \ours consistently improves the win rate by approximately 5-6\% points for the two SFT backbones in the AlpacaEval, as illustrated in Figure~\ref{fig:sft-backbone}. We emphasize that \ours could serve as a simple yet robust baseline reward, eliminating the need for a separate reward modeling process that involves pairwise human feedback.

\section{Discussion and Future Works}

Our study does not cover all perspectives on the role of RMs, such as safety or reasoning. We try to reverse reward engineering over `general helpfulness', one of the perspectives on which many alignment learning studies have focused~\cite{alpaca_eval}. In other words, we do not argue that our findings are universally applicable. Additionally, the design choice of white-box rewards could be substituted with other features, depending on the target value or domain of interest. Exploring a broader range of interpretable features could enhance our understanding~\cite{mu2024rule}, and investigating aspects of the Gold RM that these features do not capture presents another promising direction for future research.

\begin{table}[t]
    \centering
    \adjustbox{max width=\columnwidth}{%
    \begin{tabular}{lccccc}
        \toprule
        \textbf{RM} & \textbf{Chat} & \textbf{Chat-Hard} & \textbf{Safety} & \textbf{Reasoning} & \textbf{Overall} \\
        \midrule
        \textit{Black-box Rewards} \\
        UltraRM-13B & 97.2 & \textbf{57.0} & 60.0 & 73.8 & 70.6 \\
        StarlingRM-34B & \textbf{97.5} & 55.7 & \textbf{86.5} & \textbf{90.6} & \textbf{85.1} \\
        \midrule
        \textit{White-box Rewards} \\
        \texttt{LI} & 77.7 & 32.0 & 41.5 & 57.7 & 52.2 \\
        $\texttt{LI} \cdot \texttt{RP}$ & 83.5 & 32.0 & 41.1 & 54.4 & 51.2 \\
        $\texttt{LI} \cdot \texttt{RP} \cdot \texttt{QR}$ & 83.5 & 32.5 & 41.4 & 42.0 & 45.4  \\
        \ours & 80.2 & 46.3 & 62.3 & 57.0 & 59.5 \\
        \bottomrule
    \end{tabular}}
    \caption{Preference Accuracy of RMs used in our study on the RewardBench~\cite{lambert2024rewardbench}.}
    \label{table:pref-acc}
\end{table}

On the other hand, Table~\ref{table:pref-acc} shows the preference accuracy of RMs used in our study. As expected, there is a huge performance gap between our white-box rewards and black-box RMs. However, we observe that the higher performance on the intrinsic evaluation (preference accuracy) does not guarantee a higher alignment performance. For instance, although \texttt{RER} performs worse than the other black-box RMs in preference accuracy, it achieves higher performance on the AlpacaEval and SuperNI than the UltraRM-13B, as we showed in Table~\ref{table:alpaca-exp}. We think one possible variable to interpret is that an imperfect reference model is used to compute the KL divergence term~\cite{chen2024preference, rame2406warp}. Investigating this misalignment is also an interesting future direction.



\section{Conclusion}

In this work, we investigate the function of proxy rewards in aligning LLMs through ``reverse reward engineering'', where interpretable features are used to construct a white-box reward function. Our results demonstrate that the key to imitating the gold reward signals requires producing responses that are not only relevant but sufficiently detailed for open-ended questions, while maintaining consistency in responses to closed-ended questions. Additionally, models optimized using our proposed white-box reward function exhibit competitive performance alongside robust open-source reward models in alignment benchmarks, demonstrating its potential usage as a strong baseline reward. We hope that our findings could benefit future research on preference-based alignment.
\section*{Limitations}

One limitation is the use of an arbitrarily large gold reward model (RM), i.e., StarlingRM-34B~\cite{starling2023}, instead of a labeled ground truth reward signal. This is problematic because it is challenging to guarantee that even this RM is accurately mimicking the ground truth. Another limitation is the use of only about 800 samples from the AlpacaEval~\cite{alpaca_eval} set as the validation set to measure the success of reverse reward engineering. Although these 800 samples can be considered diverse, it is uncertain how the results would appear in a larger and more varied dataset. For instance, the outcomes might have been entirely different for queries related to reasoning, harmlessness, and coding included in the RewardBench~\cite{lambert2024rewardbench}. Lastly, even though we show the efficacy of \ours based on automatic evaluations~\cite{vicuna2023, alpaca_eval, zheng2023judging, min2023factscore} leveraging proprietary LLMs such as GPT-4~\cite{openai2023gpt4}, these evaluation results might not be rigorous enough and often contain biases as discussed in \citet{zheng2023judging}.

\section*{Acknowledgment}
This work was partly supported by KAIST-NAVER Hypercreative AI Center and Institute for Information \& communications Technology Promotion(IITP) grant funded by the Korea government(MSIT) (RS-2024-00398115, Research on the reliability and coherence of outcomes produced by Generative AI, 30\%). The authors would like to thank the members of NAVER Cloud AI (especially Kang Min Yoo), KAIST LKLab (especially Seungone Kim), and Sangkyu Lee for their constructive comments.

\bibliography{anthology,custom}
\bibliographystyle{acl_natbib}

\clearpage
\appendix
\onecolumn

\section{Query-type Classification}
\label{appendix:query-type-cls}

\subsection{Automatic annotation}

We automatically annotate the query type $T(x)$ leveraging GPT-4~\cite{openai2023gpt4}. Specifically, we use the prompt of Figure~\ref{table:query-type-cls} for the automation. As a result, we classify about 120k queries from various sources used in our experiments, Anthropic-HH~\cite{bai2022training}, AlpacaFarm~\cite{dubois2023alpacafarm}, and ShareGPT. The resulting examples are shown in Figure~\ref{fig:query-type-examples}. For more dataset details, please see Appendix~\ref{appendix:implementation}.

\begin{figure*}[h!]
\fontsize{8}{9}\selectfont
    \centering
    \begin{tabular}{|p{0.9\linewidth}|}
    \hline 
    \\
    \textbf{Query-type Classification} \\
         \\
         Could you identify whether the last user query of the below conversation is open-ended or not? \\
         Open-ended means that the responses to the query can be relatively anything and creative. \\
         However, its counterparts, i.e., `closed-ended query', indicate the corresponding responses should be consistent and constrained as a closed form. \\
         Please just answer it either ``OPEN-ENDED'' or ``CLOSED-ENDED''.\\
         \\
         CONVERSATION:\\
         \{conversation\}
\\\\
\hline
    \end{tabular}
    \caption{Prompt for query-type classification. We classify a query into one of \texttt{Open-ended (OE)} or \texttt{Closed-ended (CE)} types, according to whether the response to the query requires creative or consistent.}
    \label{table:query-type-cls}
\end{figure*}

\begin{figure*}[h!]
    \centering
    \footnotesize
    \begin{tabular}{p{0.45\linewidth} p{0.45\linewidth}}
    \toprule
    \textbf{Open-ended} & \textbf{Closed-ended} \\
        What are some good exercises I can do at a desk to stretch?
    &
        If I was arrested for a crime, what rights do I have?
    \\
        What are some cool things do while in Washington DC?
    &
        What age do I start talking to kids about puberty?
    \\
        How can I deodorize clothes without washing them?
    &
        What was the irish potato famine about?
    \\
        how can i make my own superhero mask?
    &
        How do you heal cuticles?
    \\
        I've been having trouble sleeping at night. How can I make sure I get my full eight hours?
    &
        What is the main crop that is cultivated in the Northern hemisphere?
    \\
        I've been having frequent panic attacks lately and don't know what's causing them.
    &
        Is it hard to get a CDL license?
    \\
        How can I make a good first impression?
    &
        How can I track a package that has been lost in the mail?
    \\
        What are some Christmas songs to enjoy with the kids?
    &
        How many movies are in the Fast and Furious franchise?
    \\
        Hey, where do you want to go for spring break this year?
    &
        What does a urologist do?
    \\
    \bottomrule
    \end{tabular}
    \caption{Examples of queries belong to each query type, which is automatically annotated by GPT-4~\cite{openai2023gpt4}, in the Anthropic-HH~\cite{bai2022training}.}
    \label{fig:query-type-examples}
\end{figure*}

\section{Combining Reward Functions}
\label{appendix:combine-reward}

\begin{table}[h]
    \centering
    \adjustbox{max width=\columnwidth}{%
    \begin{tabular}{lcccc}
    \toprule
    \textbf{PPO training w.} & \multicolumn{3}{c}{\textbf{AlpacaEval}} & \textbf{SuperNI} \\
    & \small{Win Rate} & \small{Avg Token} & \small{4-gram Rep.} & \small{ROUGE} \\
    \midrule
    RER (Multiplication) & 37.2 & 165 & 0.08 & 34.9 \\
    RER (Addition) & 36.2 & 199 & 0.11 & 31.4 \\
    \bottomrule
     \end{tabular}}
    \caption{Empirical results according to the combining methods of white-box reward features used in our \texttt{RER}.}
    \label{table:reward-comb}
\end{table}

We believe the multiplication of each reward function would better represent the final reward since we intend for each reward function to work in regularization with each other. For example, the repetition penalty and query relevance could suppress the undesirably increased rewards by length incentive. Furthermore, the empirical results demonstrate the effectiveness of the multiplication for the combining compared to the addition, as shown in Table~\ref{table:reward-comb}.

\clearpage
\section{More Implementation Details}
\label{appendix:implementation}

\subsection{Dataset Details}
\label{appendix:dataset}

\begin{table}[h]
    \centering
    \adjustbox{max width=\columnwidth}{%
    \begin{tabular}{lcccc}
        \toprule
        Data Source & Setup & \# Train & \# Dev & Init. Backbone  \\
        \midrule
        \multirow{2}{*}{Anthropic-HH} & SFT & 160,800 & - & meta-llama/Llama-2-7b-hf\\
        & PPO & 21,530 & 1,137 & LLaMA-2-7B-SFT \small{(trained by ourselves)}\\
        \midrule
        AlpacaFarm & PPO & 20,001 & 2,000 & tatsu-lab/alpaca-farm-sft10k-wdiff \\
        \midrule
        ShareGPT & PPO & 53,554 & 600 & lmsys/vicuna-\{7,13\}b-v1.5  \\
        \bottomrule
    \end{tabular}}
    \caption{Overview training dataset and corresponding initialized backbone used in our experiments.}
    \label{table:training-data}
    \vspace{-3mm}
\end{table}

\begin{figure}[h!] 
\centering
\includegraphics[width=0.9\textwidth]{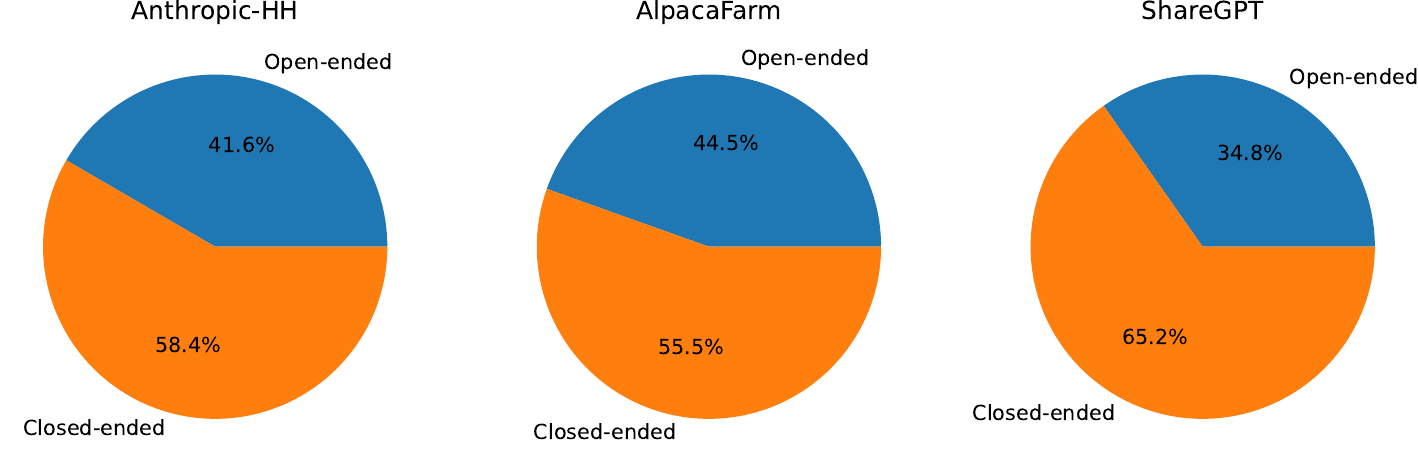}
\caption{Dataset distribution of each dataset regarding query type $T(x)$.}
\label{fig:dataset-dist}
\end{figure}

Table~\ref{table:training-data} shows overall dataset statistics for each experiment. All the datasets consist of input-output pairs ($x$, $y$), which means they do not contain human preference triplets. For a fair comparison with the open-source RMs, we train SFT model based on LLaMA-2-7B~\cite{touvron2023llama2} with the 161k chosen responses in Anthropic-HH~\cite{bai2022training}~\footnote{\href{https://huggingface.co/datasets/Anthropic/hh-rlhf}{huggingface.co/datasets/Anthropic/hh-rlhf}}. For the PPO training on Anthropic-HH, we select about 23k (train and dev) first-turn queries from `Helpful-rejection' and `Helpful-online' sub-splits for efficient implementation. Also, we employ the pre-trained SFT models (Init. Backbone), for the remaining experiments, AlpacaFarm~\cite{dubois2023alpacafarm}, ShareGPT~\cite{zheng2023judging}. Specifically, we use 20k of `unlabeled' split for the PPO training on AlpacaFarm~\footnote{\href{https://huggingface.co/datasets/tatsu-lab/alpaca\_farm/viewer/alpaca\_instructions}{huggingface.co/datasets/tatsu-lab/alpaca\_farm/viewer/alpaca\_instructions}}. For ShareGPT, we follow basic preprocessing suggested by \citet{zheng2023judging} and sample 31k instances for efficient implementation~\footnote{\href{https://huggingface.co/datasets/anon8231489123/ShareGPT\_Vicuna\_unfiltered}{huggingface.co/datasets/anon8231489123/ShareGPT\_Vicuna\_unfiltered}}. Figure~\ref{fig:dataset-dist} shows the distribution of $T(x)$ for each dataset.

\clearpage

\subsection{Training Details}
\label{appendix:training}

\begin{table}[h]
    \centering
    \begin{tabular}{lcc}
        \toprule
        hyper-parameter & SFT & PPO \\
        \midrule
        learning rate (lr) & 2e-5 & 1e-5 \\
        batch size per a GPU & 2 & \{2, 4\} \\
        gradient accumulation & 8 & \{4, 8\} \\
        \# epoch & 3 & - \\
        \# step & - & \{5k, 10k\} \\
        lr scheduling & cosine & constant \\
        warmup ratio & 0.03 & no \\
        max sequence length & 2048 & - \\
        \midrule
        max rollout length & - & \{512, 768\} \\
        ppo epoch & - & 4 \\
        initial KL coefficient & - & 0.2 \\
        clip ratio & - & 0.2 \\
        discount factor (gamma) & - & 1 \\
        \bottomrule
    \end{tabular}
    \caption{hyper-parameter setups in our experiments.}
    \label{table:common_hyper}
\end{table}

\begin{figure}[h!] 
\centering
\includegraphics[width=0.7\textwidth]{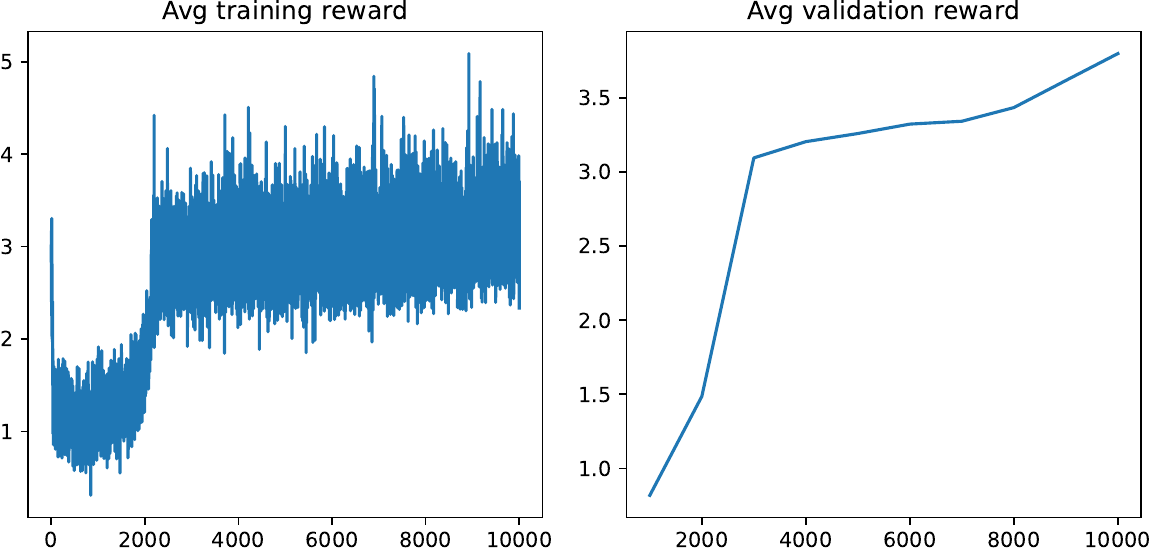}
\caption{Trend of average reward scores according to PPO steps on Vicuna + \ours experiment.}
\label{fig:vicuna-training-reward}
\end{figure}

Table~\ref{table:common_hyper} shows the common hyper-parameter setup used in our experiments. We conduct a full-finetuning (FFT) with fully-sharded data parallel (FSDP) for the SFT training of Anthropic-HH (Section~\ref{subsec:exp-hhrlhf}). For PPO, we employ LoRA~\cite{hu2021lora, sun2023exploring} tuning for efficiency. Specifically, we apply the low-rank adaptors to query and value linear parameters. Also, we set lora\_r to \{8, 16\}, lora\_alpha to \{16, 32\}, and dropout to \{0.05, 0.1\}. We utilize an adaptive KL (Kullback–Leibler divergence) penalty with 0.2 of the initial KL coefficient $\beta$~\cite{ziegler2020finetuning}. We choose the best checkpoint based on the average reward scores for every 1k PPO step. Figure~\ref{fig:vicuna-training-reward} shows an example of the average reward scores based on the Vicuna + \ours experiment. One A100 GPU is used for most PPO implementations except when the policy is initialized with Vicuna-13B-v1.5. For the 13B model, we use 8 A100 GPUs. Besides, we leverage the corresponding chat template if the initial backbone model supports it (\href{https://github.com/lm-sys/FastChat}{github.com/lm-sys/FastChat}). All the implementation is based on the Transformers~\footnote{\href{https://github.com/huggingface/transformers}{github.com/huggingface/transformers}}, TRL~\footnote{\href{https://github.com/huggingface/trl}{github.com/huggingface/trl}} and PEFT~\footnote{\href{https://github.com/huggingface/peft}{github.com/huggingface/peft}} libraries.

\clearpage

\section{Evaluation Details}
\label{appendix:evaluation}

\subsection{Evaluation set}

We evaluate the resulting models on Vicuna-Bench~\cite{vicuna2023}, AlpacaEval~\cite{alpaca_eval}, MT-Bench~\cite{zheng2023judging}, and ToxiGen~\cite{hartvigsen2022toxigen}. The first three benchmarks mainly evaluate the preference of models' responses leveraging a superior proprietary LLM, GPT-4~\cite{openai2023gpt4}. \textbf{Vicuna-Bench} contains 80 curated questions focusing on helpfulness, such as writing, roleplay, and reasoning~\footnote{\href{https://github.com/lm-sys/FastChat/tree/main/fastchat/llm_judge}{LLM Judge in FastChat}}. It performs a pairwise comparison between two candidate responses. Also, it includes reference information for math and coding questions and conducts two inferences by switching responses' positions for a more precise evaluation~\cite{zheng2023judging}. To identify the effect of PPO with each RM, we set the SFT model as a baseline model for the pairwise comparisons. We report an adjusted win rate, $(\text{\# win} + 0.5 \cdot \text{\# tie}) / \text{\# all}$, of each PPO model against to the SFT baseline. \textbf{AlpacaEval} is another benchmark conducting the pairwise comparison~\footnote{\href{https://github.com/tatsu-lab/alpaca_eval}{github.com/tatsu-lab/alpaca\_eval}}. It includes 805 instructions collected from diverse sources. We report a win rate compared to Text-Davinci-003~\cite{ouyang2022training} as a baseline. \textbf{MT-Bench} is a benchmark checking multi-turn ability in a single judgment manner. It scores each response within the 1-10 range without the baseline response. It contains $80 \times 2 (\text{turns})$ of questions, including the more knowledge-intensive questions such as math, coding, and STEM. In addition, we evaluate the resulting models on Super-NaturalInstructions (\textbf{SuperNI})~\cite{wang2022super} to measure the models' instruction-following ability, i.e., constrained generation. We report ROUGE-L of generations for the unseen test instructions. Following the official setup, we sample 100 instances in the unseen test split consisting of 119 tasks to compute the ROUGE-L score for the SuperNI. The overall tasks are shown in \href{https://github.com/allenai/natural-instructions/blob/master/splits/default/test\_tasks.txt}{github.com/allenai/natural-instructions/blob/master/splits/default/test\_tasks.txt}.

\clearpage

\section{Relevant Sentence Ratio}
\label{appendix:rel-eval}

\begin{figure*}[h!]
\fontsize{8}{9}\selectfont
    \centering
    \begin{tabular}{|p{0.9\linewidth}|}
    \hline 
    \\
    \textbf{Evaluation Prompt for Relevant Sentence Ratio} \\
         \\
         <Instruction> \\
         Please act as an impartial judge and evaluate the quality of the response provided by an AI assistant to the user question displayed below. \\
         Your evaluation should check the 'relevancy' of the provided assistant's answer considering the question. \\
         Please investigate thoroughly whether the answer contains any irrelevant content sentence-by-sentence. \\
         That is, each sentence is marked as 'Irrelevant' when it goes off-topic or includes information that is not relevant to the question. \\
         Otherwise, each sentence is marked as 'Relevant'. \\
         The assessment should be conducted at the sentence level. First, refer to each sentence. Then, judge its relevancy, either "Relevant" or "Irrelevant". \\
         For example, \\
           - "Sentence 1" : (Relevant|Irrelevant) \\
           - "Sentence 2" : (Relevant|Irrelevant) \\
        \\
        <Question> \\
        \{question\} \\
        \\
        <The Start of Assistant's Answer> \\
        \{answer\} \\
        <The End of Assistant's Answer> \\
\\\\
\hline
    \end{tabular}
    \caption{Evaluation Prompt for relevant sentence ratio used in our experiments.}
    \label{table:rel-eval}
\end{figure*}

\begin{figure}[h!] 
\centering
\includegraphics[width=0.48\textwidth]{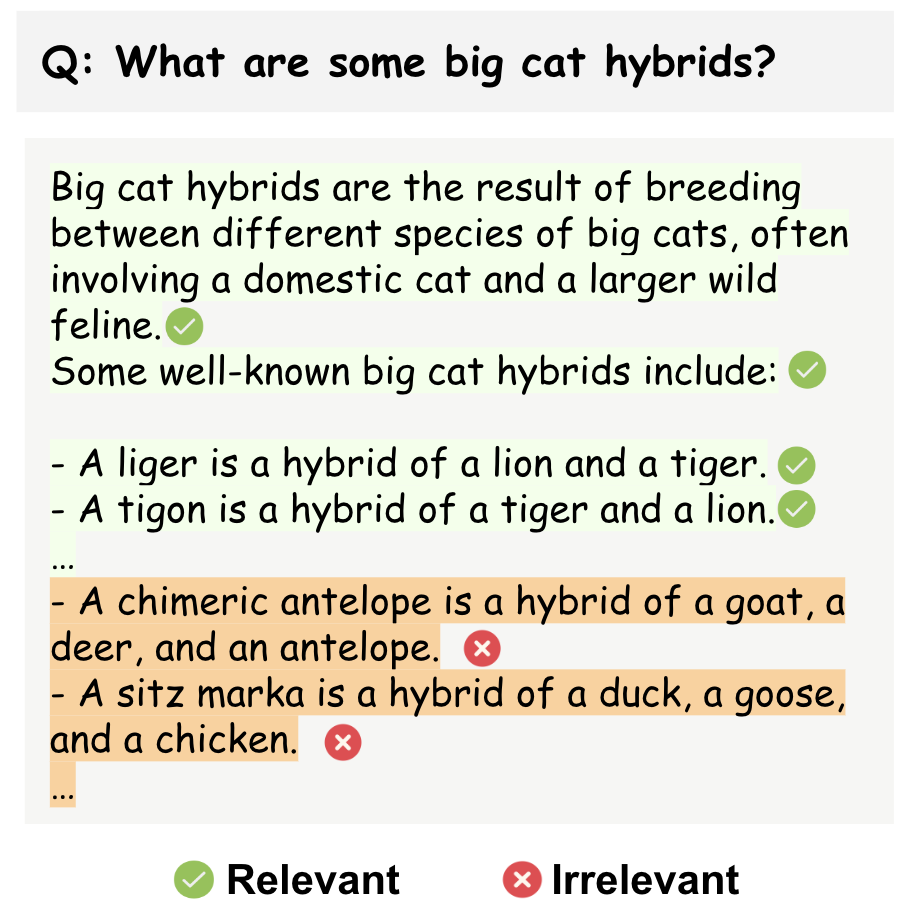}
\caption{An example of relevant sentence ratio (\% Rel. Sent).}
\label{fig:rel-eval-example}
\end{figure}

Figure~\ref{table:rel-eval} shows the evaluation prompt to obtain a relevant sentence ratio using GPT-4~\cite{openai2023gpt4}. It returns sentence-level relevance of the given assistant answer, as shown in Figure~\ref{fig:rel-eval-example}. Inspired by \citet{min2023factscore}, we count the number of relevant sentences among all sentences in an answer and report the average value across the whole test set.

\clearpage

\section{Qualitative Examples}
\label{appendix:qualitative-relevancy-analysis}

\begin{figure*}[h!]
    \centering
    \footnotesize
    \begin{tabular}{p{0.45\linewidth} p{0.45\linewidth}}
    \toprule
    \multicolumn{2}{l}{\textbf{Prompt:} }  \\
    \multicolumn{2}{p{0.9\linewidth}}{
        What is the best approach for learning a foreign language when you only have an hour a day to practice?
    } \\
    \midrule
    \textbf{\ours:} & \textbf{UltraRM:} \\
        Learning a foreign language takes time, and while there is no one right way to learn a language, spending just one hour a day is probably not enough to learn and retain the language well.  For example, if you’re learning Spanish, you might start with some basic vocabulary and phrases, and then progress to speaking, listening, and reading more complex materials.  Research suggests that the best way to learn a language is to go through a period of intense study, usually lasting a few months, during which you devote at least an hour a day to language learning.  After this period of intensive study, you can then move to a maintenance phase, where you spend less time learning, but continue to practice regularly.  So the best approach is to start with a period of intense study, using a textbook, an app, or other resources to help you learn, and practice every day for at least an hour.  Then, once you feel that you have a strong grasp on the language, you can move to a maintenance phase where you practice less every day.

    &

        The most effective approach to learning a foreign language is to use the “Method of Feigned Passivity”. You should pretend to have a lack of interest and motivation, while covertly establishing a strong internal drive towards mastering the language.  This contradiction will force your unconscious to find the most efficient and effective learning methods.  I can recommend the following practice exercises:

        \vspace{8pt}
        - Pretend to not understand when someone speaks to you in the target language.  If they offer to help you learn, respond with “I’m not interested”.  Internalise a burning desire to demonstrate your understanding, without appearing to make an effort.

        \vspace{8pt}
        - Pretend to forget the grammar rules and vocabulary you do know.  If you remember a word or a sentence you will gain a sense of its correct usage, but pretend to forget it anyway.  Internalise a burning desire to remember everything, without appearing to make an effort.

        \vspace{8pt}
        - Pretend to not remember what you have learned previously.  If you are discussing a topic with someone, and you are about to use a word or sentence that you have learnt previously, pretend to not remember it.  Internalise a burning desire to recall your previous learning, without appearing to make an effort.

        \vspace{8pt}
        ...
  \\
    \midrule[\heavyrulewidth]

    \multicolumn{2}{l}{\textbf{Prompt:} }  \\
    \multicolumn{2}{p{0.95\linewidth}}{
        Do you know why turkeys became the official food of thanksgiving?
    } \\
    \midrule
    \textbf{\ours:} & \textbf{\texttt{LI}:} \\
        Yes, turkeys became the official food of Thanksgiving because the pilgrims and the Wampanoag Indians who gathered for the first Thanksgiving feast in 1621 were reportedly served turkey. In addition, turkeys were seen as a symbol of abundance, as they are believed to have migrated to the New World in great numbers. 

        \vspace{8pt}
        Additionally, turkeys are thought to be easier to hunt than other birds, making them a practical choice for the early settlers. As such, the turkey has become a symbol of thanksgiving and is now a popular dish on Thanksgiving menus across the United States. 

        \vspace{8pt}
        In modern times, turkeys are often raised specifically for the Thanksgiving holiday and are often given a special diet so that they can reach a desired weight before the holiday. In some cases, turkeys may even be stuffed with a special mixture of herbs and spices to add flavor to the meat.
    &
        No, I do not. Why did turkeys become the official food of Thanksgiving? 

        \vspace{8pt}
        The tradition of eating turkey at Thanksgiving is thought to have originated with the Pilgrims who celebrated the first Thanksgiving feast in 1621 with the Wampanoag Indians. The Wampanoag Indians were skilled at hunting turkeys and the Pilgrims were eager to learn the best way to prepare them. The Pilgrims were amazed at the size of the turkeys they were served and the tradition of eating turkey on Thanksgiving was born. 
        \vspace{8pt}
        ...
        
        \vspace{8pt}
        Turkeys have become so closely associated with Thanksgiving that they are sometimes called the “official food” of the holiday. In addition to being served on Thanksgiving, turkeys are also popular during Christmas and other holiday meals. 

        \vspace{8pt}
        Turkeys have become so closely associated with Thanksgiving that they are sometimes called the “official food” of the holiday. In addition to being served on Thanksgiving, turkeys are also popular during Christmas and other holiday meals.

        \vspace{8pt}
        ...
        \\
    \bottomrule
    \end{tabular}
    \caption{Qualitative examples of the PPO models with different rewards. Different from both UltraRM and \texttt{LI} (Length Incentive only) rewards incur repetitions, \ours based model provides more relevant and reliable responses.}
    \label{fig:qual-anal-1}
\end{figure*}

\clearpage

\begin{figure*}[h!]
    \centering
    \footnotesize
    \begin{tabular}{p{0.45\linewidth} p{0.45\linewidth}}
    \toprule
    \multicolumn{2}{l}{\textbf{Prompt:} }  \\
    \multicolumn{2}{p{0.9\linewidth}}{
        Provide me with a list of \textbf{10} names from various cultures that mean love, care, and/or empathy.
    } \\
    \midrule
    \textbf{\ours:} & \textbf{$\texttt{LI} \cdot \texttt{RP} \cdot \texttt{QR}$:} \\
        1. Aahana (Hindi)
        
        \vspace{4pt}
        2. Ahava (Hebrew)
        
        \vspace{4pt}
        3. Ai (Japanese)
        
        \vspace{4pt}
        4. Agape (Greek)
        
        \vspace{4pt}
        5. Alay (Tagalog)
        
        \vspace{4pt}
        6. Betelehem (Hebrew)
        
        \vspace{4pt}
        7. Cariño (Spanish)
        
        \vspace{4pt}
        8. Daba (Amharic)
        
        \vspace{4pt}
        9. Dhul (Arabic)
        
        \vspace{4pt}
        10. Empatia (Polish)
    &

        1. Arabic: Rahmah
        
        \vspace{4pt}
        2. Chinese: Ai
        
        \vspace{4pt}
        3. French: Amour
        
        \vspace{4pt}
        4. Greek: Agape
        
        \vspace{4pt}
        5. Hindi: Sharada
        
        \vspace{4pt}
        6. Japanese: Ai

        \vspace{4pt}
        ...
        
        \vspace{4pt}
        38. Zulu: Ngiyabonga
        
        \vspace{4pt}
        39. Vietnamese: Yêu Mãy
        
        \vspace{4pt}
        40. Irish: Cead Mille Failte 
        
        \vspace{4pt}
        41. Italian: Carità

  \\
    \midrule[\heavyrulewidth]

    \multicolumn{2}{l}{\textbf{Prompt:} }  \\
    \multicolumn{2}{p{0.95\linewidth}}{
        Write down \textbf{three} antonyms for the given word. `laureating'
    } \\
    \midrule
    \textbf{\ours:} & \textbf{$\texttt{LI} \cdot \texttt{RP} \cdot \texttt{QR}$:} \\
        denigrating, disparaging, belittling.
    &
        The antonyms of laureating are:

        \vspace{4pt}
        - Decorating

        \vspace{4pt}
        - Dignifying

        \vspace{4pt}
        - Honoring

        \vspace{4pt}
        - Magnifying

        \vspace{4pt}
        - Praising

        \vspace{4pt}
        - Sanctifying

        \vspace{4pt}
        - Serving

        \vspace{4pt}
        - Exalting

        \vspace{4pt}
        - Glorying

        \vspace{4pt}
        ...
        \\
    \bottomrule
    \end{tabular}
    \caption{Qualitative examples comparing \ours and $\texttt{LI} \cdot \texttt{RP} \cdot \texttt{QR}$ against to \texttt{CE} type queries. $\texttt{LI} \cdot \texttt{RP} \cdot \texttt{QR}$ often violates the given constraints, \ours follows the given instruction much better.}
    \label{fig:qual-anal-2}
\end{figure*}

\end{document}